\RequirePackage{etoolbox}
\csdef{input@path}%
{%
 {sty/},
 {img/},
}
\documentclass{elsevierbook}

\let\AfterPackage\undefined
\usepackage[square,sort,comma,numbers]{natbib}
\usepackage{amssymb}
\usepackage{times}
\usepackage{epsfig}
\usepackage{graphicx}
\usepackage{amsmath}
\usepackage{amssymb}
\usepackage{color}
\usepackage{parskip}
\usepackage{url}
\usepackage{nomencl}
\usepackage{multirow}
\usepackage{hhline}
\usepackage{color}
\usepackage{calc}
\usepackage{booktabs}

\begin{document}


\Mainmatter
  \begin{frontmatter}

\chapter{Multimodal perception for dexterous manipulation}\label{chap1}
\begin{aug}
\author[addressrefs={ad1}]%
{%
\fnm{Guanqun} \snm{Cao}%
}%
\author[addressrefs={ad1}]%
{%
\fnm{Shan} \snm{Luo}%
}%
\address[id=ad1]%
{%
smARTLab,
Department of Computer Science,
University of Liverpool,
United Kingdom. \\
Emails: {g.cao, shan.luo}@liverpool.ac.uk.
}%

\end{aug}
\begin{abstract}
Humans usually perceive the world in a multimodal way that vision, touch, sound are utilised to understand surroundings from various dimensions. 
These senses are combined together to achieve a synergistic effect where the learning is more effectively than using each sense separately.
For robotics, vision and touch are two key senses for the dexterous manipulation.
Vision usually gives us apparent features like shape, color, and the touch provides local information such as friction, texture, etc. 
Due to the complementary properties between visual and tactile senses, it is desirable for us to combine vision and touch for a synergistic perception and manipulation.
Many researches have been investigated about multimodal perception such as cross-modal learning, 3D reconstruction, multimodal translation with vision and touch. 
Specifically, we propose a cross-modal sensory data generation framework for the translation between vision and  touch, which is able to generate realistic pseudo data.
By using this cross-modal translation method, it is desirable for us to make up inaccessible data, helping us to learn the object's properties from different views. 
Recently, the attention mechanism becomes a popular method either in visual perception or in tactile perception. 
We propose a spatio-temporal attention model for tactile texture recognition, which takes both spatial features and time dimension into consideration.
Our proposed method not only pays attention to the salient features in each spatial feature, but also models the temporal correlation in the through the time. 
The obvious improvement proves the efficiency of our selective attention mechanism.
The spatio-temporal attention method has potential in many applications such as grasping, recognition, and multimodal perception.
\end{abstract}

\begin{keywords}
\kwd{Multimodal Perception}
\kwd{Vision}
\kwd{Touch}
\kwd{Multimodal Translation}
\kwd{Attention Mechanism}
\end{keywords}

\end{frontmatter}

\section{Introduction}
\label{sec:sample1}
As in humans, even in the daily task such as cleaning teeth, both visual and tactile senses are implemented subconsciously.
People use eyes to recognise and locate the toothbrush. 
When the toothbrush is in the mouth, people use tactile feedback to complete the blind spot cleaning due to the occluded views.
Humans interact and perceive the world with various feelings including the vision, touch, sound, taste, and smell. 
Multiple feelings can be combined together to have a synergistic effect where the results can not be accomplished by using each sense separately.
 
For robotic manipulation, vision and touch are two main sensory modalities for perception.
In robotics, vision is often obtained with a camera by capturing the image from a distance, and it can provide us the appearance, shape, and color of the object. 
The visual modality has been extensively implemented in many applications such as recognition \cite{he2016deep}, object detection \cite{redmon2017yolo9000}, and pose estimation \cite{brachmann2014learning}. 
However, it is still challenging to complete the manipulation task using a single visual modality.
For example, it is difficult for robots to recognise the transparent objects that the color of which is same to surroundings.
In addition to vision, the tactile information is also used by robotic, enabling the understanding of the world from another dimension.
The tactile information is achieved by direct interaction between tactile sensors and contacted objects, which can give us the object's surface properties including the textures, local geometry, friction, hardness even under occlusions.
The transparent object, that can not be detected by vision, is able to be easily recognised through touching. 
Recently, researchers have achieved considerable progress on the tactile sensing \cite{luo2017robotic,gomes2021generation,gomes2020blocks,gomes2020geltip}, such as object recognition \cite{gorges2010haptic,kim2005texture,li2013sensing,luo2015novel,luo2016iterative,luo2015tactile,luo2019iclap}, slip detection \cite{li2018slip,james2018slip}, shape perception \cite{aggarwal2015haptic,meier2011probabilistic}, object exploration \cite{luo2015localizing} and pose estimation \cite{petrovskaya2006bayesian,bimbo2016hand}.
Due to the complementary characteristics between vision and touch, the combination of vision and touch for multimodal perception has become a popular topic \cite{kroemer2011learning,taunyazov4event,ilonen2013fusing}.

In addition, work~\cite{lacey2007vision} shows that the human brain employs a multisensory model to perceive the world in daily experience.
It indicates that the same features from various sensory modalities share a common subspace. 
Through the common subspace, it is possible to perform a unitary presentation from multiple modalities or transfer the feature from one modality to another. 
For instance, human is able to imagine the taste of food by smelling it or tell the smell by the taste, which is called synaesthesia.
Inspired by this, progressive researches have been carried out such as cross model learning especially for recognition~\cite{kroemer2011learning,taunyazov4event,falco2017cross,liu2016visual,luo2018vitac}, and multimodal translation \cite{johnson2018image,oh2019speech2face}. 
We propose a cross-modal sensory data generation framework for the translation between vision and  touch, through which we can make up realistic pseudo data to replace the inaccessible real data.

Moreover, the attention mechanism adapted from Nature Language Processing field comes to a hot topic in robotic manipulation and perception.
As the attention mechanism emphasises the salient features, it is able to learn the features more effectively and efficiently.
We propose a spatio-temporal attention model for tactile texture recognition. It not only pays focus on salient spatial features but also models the correlation of each location through the time.
The recognition results show the robustness and efficiency of our selective attention model.

The rest of paper is presented as follows:  An overview of multimodal perception is discussed including the cross-modal learning in Section \ref{2}, 3D reconstruction in Section \ref{3}, multimodal translation in Section \ref{4} and attention mechanism in Section \ref{5};
conclusions and future work are summarised in Section \ref{7}. 

\section{Cross-modal Learning}\label{2}
Vision and tactile sensing have been investigated to perceive the objects respectively.
Currently, many researchers try to match the visual information and the corresponding tactile signals.
In \cite{lacey2007vision}, it is proved that the knowledge in different modalities can be transferred. It also means different modalities shared a common latent space.
Lin \textit{et al.} \cite{lin2019learning} demonstrate the correlation between the vision and touch. The tactile features and visual features are given to recognise if or not they belong to the same object.    

Apart from matching the vision and touch, the cross-modal learning has been explored by many researchers \cite{liu2018surface,takahashi2019deep,kroemer2011learning,taunyazov4event,falco2017cross,liu2016visual,luo2018vitac}. 
In \cite{liu2018surface}, 
a cross-modal framework is developed in the visual retrieval problem. The tactile modality is used as the query to search the image in the visual modality. 
By using a structured dictionary and a classifier, the modal-invariant representation can be learned to generate corresponding visual results. 
From the opposite perspective, the visual modality is also applied to estimate the tactile properties \cite{lin2019learning}.
In this work, a shared continuous latent space is learnt between visual and tactile modalities by using an encoder-decoder network.
In \cite{kroemer2011learning,taunyazov4event,falco2017cross,liu2016visual,luo2018vitac}, they focus on the task of recognition of materials. The visual modality and corresponding tactile information are fused together, which improves the performance of recognition compared with using a single modality.
Most recently, Zheng \textit{et al.} \cite{zheng2020lifelong} propose a lifelong learning framework for material perception with vision and touch. The knowledge is accumulated in previous learning and works in later tasks, solving the inefficiency of isolated learning paradigm.

\section{3D Reconstruction}\label{3}
The vision provides us apparent characteristics such as shape, and tactile sense gives us the local geometry and precise location. 
In the object 3D reconstruction, it is difficult to use the vision to reconstruct an accurate model due to the occlusion and noise of the measurement.
The use of the tactile sensing for reconstruction is too slow due to the small contact area and the alignment is challenging.
To this end, the tactile sense, used as a complementary modality, can be applied to refine the shape of the object combined with vision.

In \cite{ilonen2013fusing}, a 3D shape of the object is firstly generated by using a single view based on point cloud and a symmetry assumption to determine the grasp point, and then the tactile sense together with vision are used to estimate the 3D shape with an Iterated Extended Kalman
Filter, minimising the uncertainty of the measurement.
The work \cite{bjorkman2013enhancing} implements the Gaussian Process regression for estimating the shape and the variance is used to measuring the uncertainty for contact. Subsequently, the shape is updated with the local contact information.
In \cite{wang20183d}, the Gaussian Process is also used for shape prediction. Differently, the learned shape priors is applied to reconstruct the shape globally and effectively rather than using a local update. To determine where to touch to refine the shape with tactile sensor, the network’s confidence score of voxel, defined by the output after sigmoid function, is proposed to measure the uncertainty of the prediction.
\section{Multimodal Translation}\label{4}
Multimodal translation is applied to transfer the data from one modality to another, which is able to provide data that are not easy to access.
Thanks to Generative Adversarial Network (GAN) \cite{goodfellow2014generative}, which is proposed to generate the data with realistic features as training data, that has been widely used in the multimodal translation \cite{johnson2018image,oh2019speech2face}. It consists of a generator and a discriminator, where the generator tries to generate new data that close to the target distribution, and the discriminator is used to classify if the input belongs to real data or generated results.

The multimodal translation based on GAN is also applied between vision and touch. 
In \cite{zhang2020generative}, a generative partial visual-tactile fused framework is proposed for the object clustering task.
The translated results are used to compensate for the missing of the real data because of occlusion and noises, which helps us to obtain inaccessible data to improve the performance of clustering.
Li \textit{et al.} \cite{li2019connecting} also propose a framework based on GAN to connect the vision and touch in robotic manipulation. The scene of how the robots interact with object is deduced using the tactile inputs as well as the tactile data of object  can be inferred with the visual scene.
What is more, the work \cite{li2019learning} applies a tactile display system to test the translated tactile signal by human. Similar to previous methods, the tactile signal is generated with the visual input, and then the signal is simulated on a vibration device so that human can test the feeling of generated signal compared with a real signal.

\subsection{“Touching to See” and “Seeing to Feel”}
\begin{table*}[t]
\centering
\begin{tabular}{lc}
\toprule
Visual-to-Tactile & Colour-SSIM \\
\midrule
Cloth Types & 0.89717 \\
Network Parameters & 0.90896 \\
ROI, with noise & 0.92059 \\
ROI, no noise & 0.91188 \\
Paired Dataset & 0.90471 \\
\bottomrule
\end{tabular}
\caption{
The similarity between visual-to-tactile images and real tactile images are measured by Colour-SSIM.
The metric is implemented on different conditions and the resulting average is calculated. }
\label{tbl:1}
\end{table*}

\begin{table*}[t]
\centering
\begin{tabular}{lc}
\toprule
Tactile-to-Visual & Colour-SSIM \\
\midrule
Cloth Types & 0.77279 \\
Network Parameters & 0.89595 \\
Paired Dataset & 0.91338 \\
\bottomrule
\end{tabular}
\caption{
The similarity between tactile-to-visual images and real visual images are measured by Colour-SSIM.
The metric is implemented on different conditions and the resulting average is calculated. }
\label{tbl:2}
\end{table*}

\begin{table*}[t]
\centering
\begin{tabular}{|c|c|c|c|c|}
\hline
& \multicolumn{2}{|c|}{Visual Images} & \multicolumn{2}{|c|}{Tactile Images} \\
\hline
Iteration & Real Images & Real/Gen. & Real Images & Real/Gen. \\
\hline
1 & 0.9623 & 0.9710 & 0.7776 & 0.8506 \\
2 & 0.9789 & 0.9770 & 0.8249 & 0.8465 \\
3 & 0.9802 & 0.9853 & 0.8902 & 0.8699 \\
4 & 0.9830 & 0.9839 & 0.8952 & 0.8736 \\
5 & 0.9858 & 0.9835 & 0.9118 & 0.8911 \\
6 & 0.9821 & 0.9867 & 0.9007 & 0.8989 \\
7 & 0.9871 & 0.9867 & 0.9127 & 0.9067 \\
8 & 0.9858 & 0.9867 & 0.9141 & 0.9044 \\
9 & 0.9876 & 0.9894 & 0.9099 & 0.8989 \\
10 & 0.9881 & 0.9894 & 0.9131 & 0.9058 \\
\hline
\end{tabular}
\caption{Classification accuracy results by using real data, and using both real data and translated data, in visual classification and tactile classification.}
\label{tbl:3}
\end{table*}
\begin{figure}
	\centering
	\includegraphics[width=0.9\columnwidth]{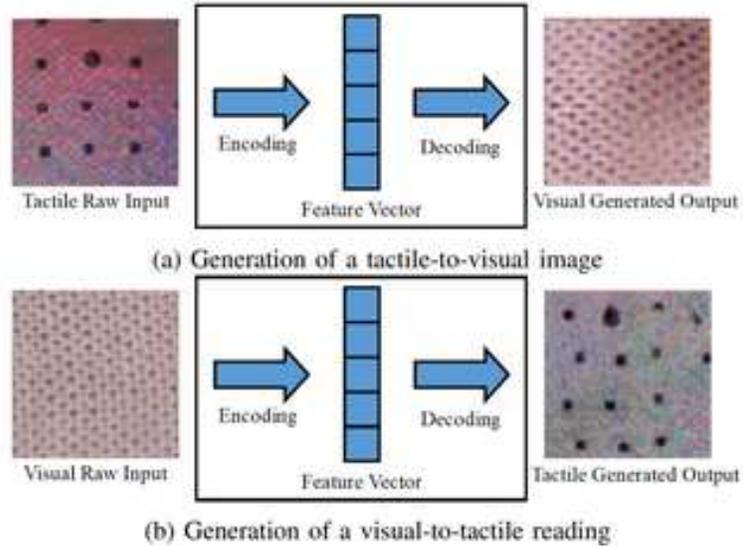}
	\caption{
     Cross-modal sensory data generation between vision and touch
	}
	\label{fig:ganframe}
\end{figure}

\begin{figure}
	\centering
	\includegraphics[width=0.9\columnwidth]{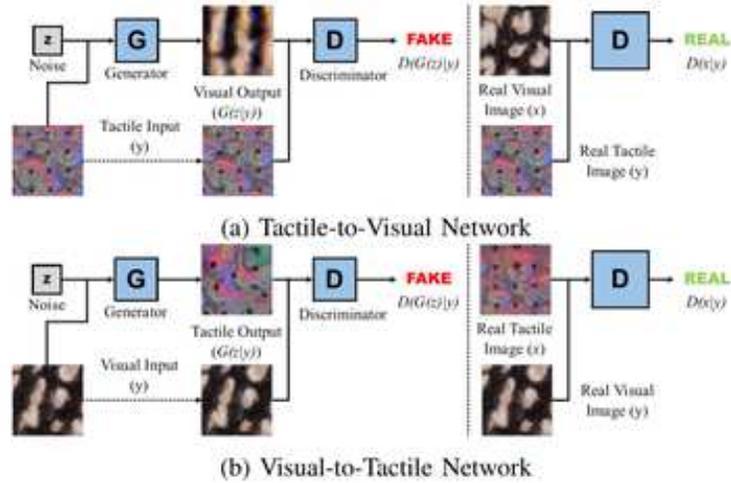}
	\caption{
     Architecture of the proposed method including visual to tactile network and tactile to visual network.
	}
	\label{fig:gandetail}
\end{figure}
Specifically, as is shown in Fig.\ref{fig:ganframe}, a framework of cross-modal sensory data generation for visual tactile perception is proposed in \cite{lee2019touching}, visual images and tactile textures from the fabric are translated to each other by using the Conditional-GAN. 
Therefore, it is possible to generate a visualisation by using tactile sense, or tactile feelings by seeing the objects, which is taken as "touching to see" and "seeing to feel".
The generated results are photorealistic compared with the real data, and the results can be used to expand the dataset to improve the performance of recognition.

\subsection{The Framework of Cross-modal Sensory Generation}
As the overall architecture is shown in Fig.\ref{fig:gandetail}, the auxiliary information $Y$ denotes data of input modality while the $X$ represents the target modality. The generator takes the $Y$ to generate the realistic pseudo data that close to the distribution of $X$, which is used to fool the fixed discriminator. The discriminator is trained to determine if the input image is from real data or generated images with a fixed generator. 
The framework includes the translation from vision to touch and touch to vision as well.
As a result, the mapping between vision and tactile modality can be learnt through this framework. 
The objective function can be represented as follows:
\begin{equation}
\begin{array}{r}
\min _{G} \max _{D} V(D, G)=\mathbb{E}_{x \sim p_{\text {data }}(x)}[\log D(x \mid y)]+ \\
\mathbb{E}_{z \sim p_{z}(z)}[\log (1-D(G(z \mid y)))]
\end{array}
\end{equation}
where the $D$ and $G$ represent the generator and discriminator respectively, $x$ and $y$ denote the data from target domain and input domain, and $z$ represents the noise used in the generation. 

\subsection{Experimental Results}
This work applies the ViTac Cloth dataset \cite{luo2018vitac}, which consists of the visual images and corresponding tactile images collected by a digital camera and a GelSight sensor \cite{yuan2015measurement} respectively from 100 pieces of fabrics. More detailed information is referred to \cite{luo2018vitac}. 
Several experiments have been conducted to measure the performance of the cross-modal translation model under different conditions. 
(1) \textit{\textbf{Cloth properties:}} Different cloth types own unique properties including the weaving patterns, colours, and textures. To this end, selected materials are used to test the performance of the model with different materials.
(2) \textit{\textbf{Network parameters:}} The internal parameters in network have different impact on the results of generated data. Therefore, the internal parameters are altered such as increasing the batch size, adding L1 loss into objective function, increasing the number of iterations, training images and resolution. 
(3) \textit{\textbf{Region of Interest:}} As the tactile data is collected from different locations of fabric, the Region of Interest (ROI) is selected to reduce the views of tactile modality that allows the training image more consistent.
(4) \textit{\textbf{Paired Dataset:}} The generated images are projected back to the original input domain, enforcing a cycle consist training. 

Moreover, a Colour Structural Similarity index (Colour-SSIM) is proposed to measure the similarity between real image and generated images, which can be represented as:
\begin{equation}
\begin{array}{l}
\text { Colour-SSIM }(x, y)= \\
\frac{\left(2 \mu_{x}^{(i)} \mu_{y}^{(i)}+C_{1}\right)\left(2 \sigma_{x y}^{(i)}+C_{2}\right)}{\left(\left(\mu_{x}^{(i)}\right)^{2}+\left(\mu_{y}^{(i)}\right)^{2}+C_{1}\right)\left(\left(\sigma_{x}^{(i)}\right)^{2}+\left(\sigma_{y}^{(i)}\right)^{2}+C_{2}\right)}
\end{array}
\end{equation}
where $\mu_{x}$ and $\mu_{y}$ represent the means of pixels value in image $x$ and $y$ respectively, $\sigma_{x}$ and $\sigma_{y}$ denotes the variance, $\sigma_{xy}$ is the covariance of $x$ and $y$, $i$ represents the index of channels in image. $C_{1}$ and $C_{2}$ are scalars with a small value.

\begin{figure}
	\centering
	\includegraphics[width=1\columnwidth]{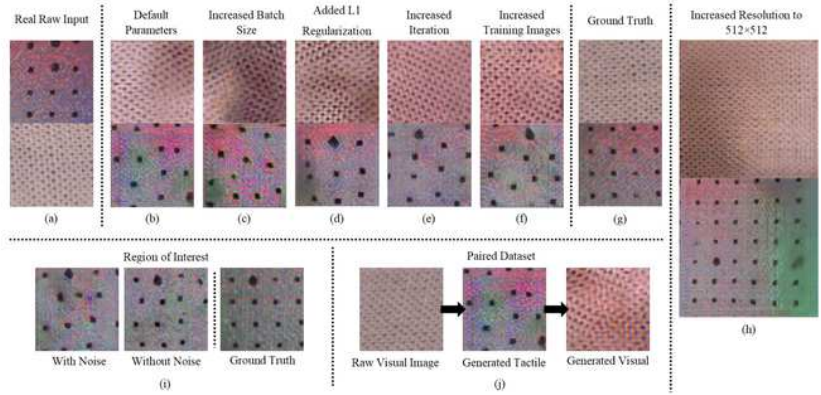}
	\caption{
     The visualisation of generated results while altering the internal parameters in the network.
     (a) Input Images (b) Outputs with default parameters (c) Increase batch size in each iteration (d) Add L1 loss to objective function (e) Increase iteration number (f) Increase amount of training images (g) Ground Truth (h) Image resolution changes from 256×256 to 512×512 (i) Select ROI (j) The output is projected back to original domain.
	}
	\label{fig:generated}
\end{figure}
From Table \ref{tbl:1}, it can be found that most methods give a result around 0.9  of Colour-SSIM as the generated tactile results own the key characteristic of the real tactile data. However, a more realistic data can be achieved with further adjusting the parameters or training methods.
From Table \ref{tbl:2}, it demonstrates a lower score compared with the results of Table \ref{tbl:1}. One reason is that the pattern and colour of visual image is difficult to be inferred from the  tactile information due to different sensing principles. In general, as can be seen from Fig.\ref{fig:generated}, the proposed methods output realistic results from both visual to tactile and tactile to visual perspective.

The generated images are also evaluated on the classification tasks. The generated results together with real data are applied on 11 categories classification compared with using real data alone.
Table \ref{tbl:3} shows that the generated data can effectively improve the performance in the early iterations.
The generated results have great potential to expand the limited dataset and enable the learning more efficiently.

\section{Attention Mechanism}\label{5}
The attention mechanism is firstly proposed in the Nature Language Processing (NLP) field \cite{bahdanau2014neural}. It is soon applied in the visual perception, such as \cite{mnih2014recurrent,hu2018squeeze,woo2018cbam}.
The attention mechanism assigns higher weights to address the salient features and lower weights to suppress the redundant features, which enables the learning effectively. 

Currently, attention mechanism becomes popular in the robotic perception and manipulation tasks.
Gu \textit{et al.} \cite{gu2019attention} propose an attention grasping network to learn the grasp points in the robotic grasping.
The salient features are paid more attention which mitigates the details lose in fully convolutional neural networks. 
In the multimodal perception of vision and touch, the vision and touch are usually fused directly that results in some redundant and in inefficient features for perception.
In \cite{cui2020self}, a new fusion method based on self-attention is used to learn representation from both modalities more effectively to predict the success of grasping. 

\subsection{Spatio-temporal Attention Model }\label{6}
\begin{figure}
	\centering
	\includegraphics[width=1.0\columnwidth]{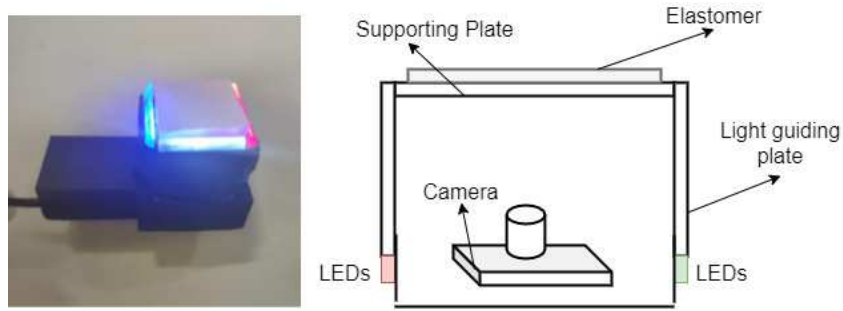}
	\caption{\textbf{\textit{GelSight Sensor.}} Left: a GelSight Sensor. Right:
	a GelSight consists of a piece elastomer coated with reflective membrane, a camera at base, LEDs and light guiding plate. When the sensor contacts the object, the elastomer is deformed due to the contact pressure. The geometry of the object's surface is taken by the deformation, which is recorded by the camera. }
	\label{fig:sensor}
\end{figure}

\begin{figure*}
	\centering
	\includegraphics[width=1\columnwidth]{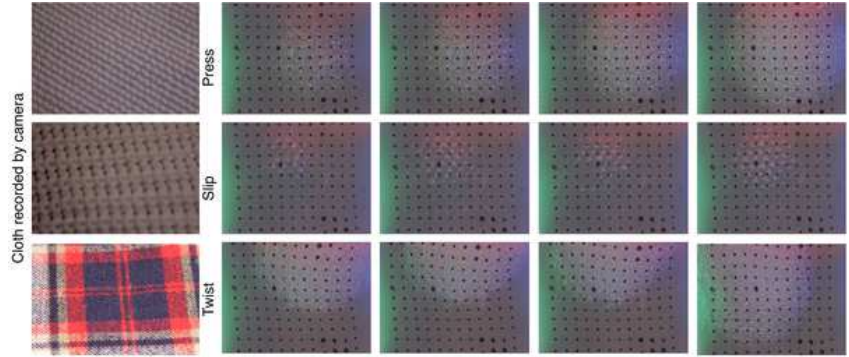}
	\caption{\textbf{\textit{Sample Tactile Sequences.}} Leftmost column: visual images recorded by a digital camera from fabrics; Right four columns: tactile sequences collected by the GelSight sensor with different interaction with fabric including pressing (first row), slipping (second row) and twisting (third row). 
	}
	\label{fig:cloth}
\end{figure*}

Recently, a spatio-temporal attention model (STAM) is proposed for the tactile texture recognition \cite{9341333}. 
The textures are shown in Fig.\ref{fig:cloth}.
The tactile textures are collected by GelSight sensor (shown in Fig.\ref{fig:sensor}) by different interaction including pressing, sipping, twisting with 100 categories fabrics.
In the spatio-temporal attention model, the spatial attention module is used to emphasise the crucial area in each frame of texture sequence, whereas temporal attention is used to select the important features considering the whole sequence regardless of the distance through the time dimension. 

\begin{figure*}[t]
	\centering
	\includegraphics[width=1\columnwidth]{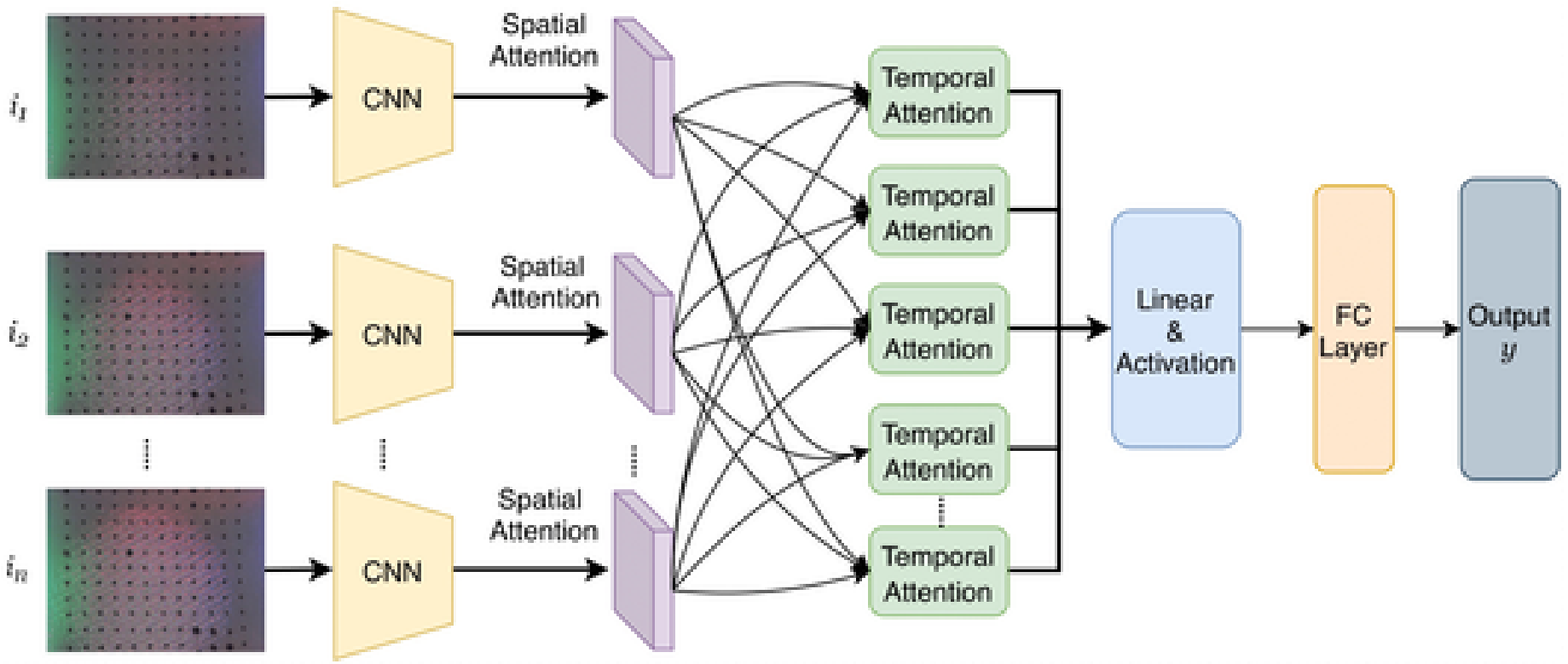}
	\caption{\textbf{\textit{The proposed STAM framework for tactile texture recognition.}} The proposed methods consists of three parts: (1) a CNN module to extract high dimensional feature map, (2) a spatial attention module to emphasise informative features in each frame, (3) a temporal attention module to model the temporal correlation in a whole sequence. 
    Finally, the model outputs a predicted category by using fully connection layers.} 
	\label{fig:framework}
\end{figure*}
The overall architecture of spatio-temporal attention model is illustrated in Fig. \ref{fig:framework}.
The texture sequence is firstly fed into a pretrained CNN to obtain a higher dimension feature map  $\boldsymbol{F} \in \mathbb{R}^{h \times w \times c}$, then the spatial attention and temporal attention are used to highlight the salient features from space dimension and time dimension respectively. At last, the network output label $y$ which is the predicted category of the texture.


\subsection{Spatial Attention}\label{8}
\begin{figure}
	\centering
	\includegraphics[width=1.0\columnwidth]{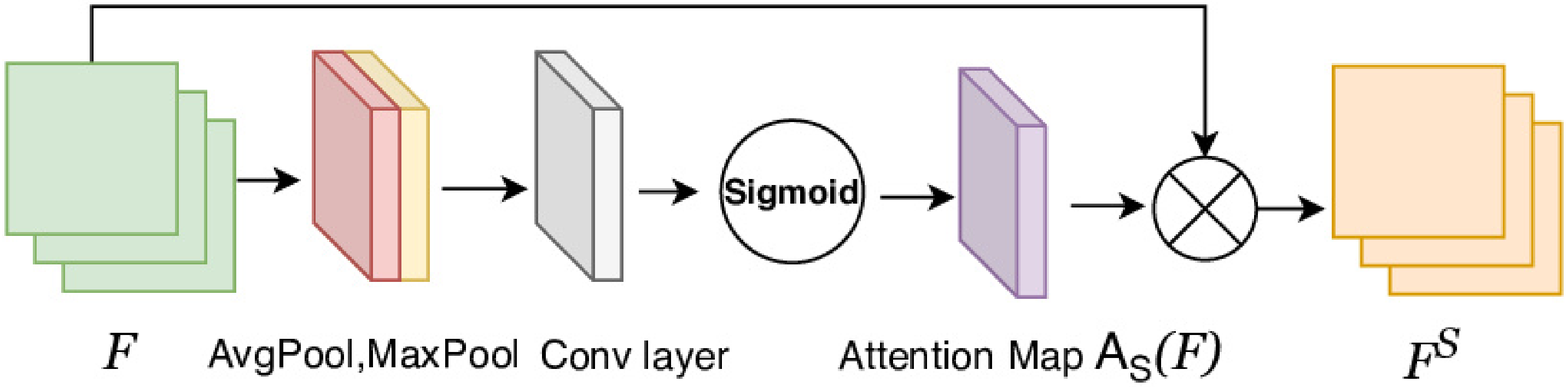}
	\caption{\textbf{\textit{Spatial Attention Module.}}  The average-pooling and max-pooling are used to learn the feature efficiently. A convolutional layer is used to calculate the attention map. } 
	\label{fig:spatial}
\end{figure}
The spatial attention aims to assign higher weights to the salient features in each frame.
As is shown in Fig.~\ref{fig:spatial}, a max-pooling and average-pooling are applied on the feature map $\boldsymbol{F}$ to describe the features effectively.
Then a convolutional layer with a kernel of $7\times7$ with a sigmoid function are used to calculate the attention map, which describes which location is more informative. 
The attention map $\boldsymbol{A_{S}(F)}$ can be calculated as:
\begin{equation} \label{pooling}
  \begin{split}
 A_{S}(F) &=\sigma (f^{7\times7}([MaxPool(F); AvgPool(F)])) \\
 &=\sigma (f^{7\times7}([F_{max}^S ; F_{avg}^S])),
 \end{split}
\end{equation}
where $ \sigma $ represents the activation function.
The output feature map can be calculated as:
\begin{equation} \label{pooling11}
  \begin{split}
F^{S} = A_{S}(F)\otimes F,
 \end{split}
\end{equation}
where $\otimes$ denotes element-wise multiplication.

\begin{figure}
	\centering
	\includegraphics[width=0.8\columnwidth]{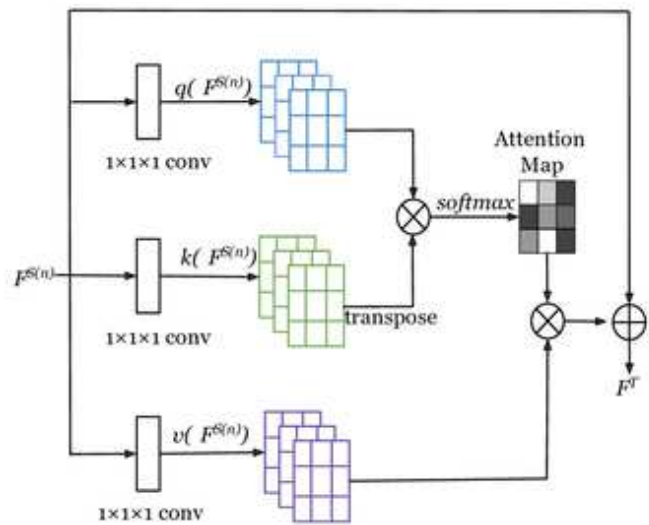}
	\caption{\textbf{\textit{Temporal Attention Module.}} 
	$1\times1\times1$ convolutions is used to turn $\boldsymbol{F^{S(n)}}$ into different feature spaces. 
     $\oplus$ denotes element-wise addition.
	}
	\label{fig:temporal}
\end{figure}
\clearpage
\subsection{Temporal Attention}\label{9}
Different from spatial attention considering a single frame, the temporal attention is used to compute the relevance of all locations in a sequence.
As is shown in Fig.\ref{fig:temporal}, each frame in the sequence is concatenated and denoted as $\boldsymbol{F^{S(n)}}$ firstly
, then $\boldsymbol{F^{S(n)}}$ is turned into to two different feature spaces by using $1\times1\times1$ convolutional layers, which can be represented as :
\begin{equation} \label{pooling1}
  \begin{split}
q(F^{S(n)}) = W_q F^{S(n)}
 \end{split}
\end{equation}
\begin{equation} \label{pooling2}
  \begin{split}
k(F^{S(n)}) = W_k F^{S(n)}
 \end{split}
\end{equation}
After, the $q(F^{S(n)})\;\text{and}\; k(F^{S(n)}) \in \mathbb{R}^{m \times c}$, where $m = {n}\times{h}\times{w}$, are used to calculate the temporal attention map that describe the correlation of all the regions regardless of time. The attention map can be computed as follows:
\begin{equation} \label{self2}
A_T(F^{S(n)})_{j, i}=\frac{\exp \left(s_{ij}\right)}{\sum_{i=1}^{m} \exp \left(s_{ij}\right)},
\end{equation}
\begin{equation} \label{self22}
\text{where}\quad s_{ij} = q\left(F^{S(n)}_{i}\right)k\left(F^{S(n)}_{j}\right)^{T} 
\end{equation}
The $A_T(F^{S(n)})_{j, i}$ illustrates the extent of $F^{S(n)}_{i}$ correlating with $F^{S(n)}_{j}$.
The temporal attention outputs the feature map: 
\begin{equation} \label{selfself}
F^{T} = {\boldsymbol(F^{T}_{1}, F^{T}_{2},...,F^{T}_{j},..., F^{T}_{m})}
\end{equation}
\begin{equation} \label{selfself2}
\text{where}\quad F^{T}_{j}=\sum_{i=1}^{m} A_T(F^{S(n)})_{j, i} v\left(F^{S(n)}_{i}\right)+F^{S(n)}_{j}
\end{equation}
\begin{equation} \label{selfself3}
v\left(F^{S(n)}\right) = W_v F^{S(n)}
\end{equation}
where $ W_v$ denotes a trainable matrix.
Moreover, as illustrated in Fig.\ref{fig:framework}, multiple temporal attention are applied which allows the model a joint learning from various feature spaces. At last, the outputs of the temporal attention module are connected with a fully connected layer to predict which category the perceived fabric is. 

\begin{table*}[htbp]
	\centering
		\caption{Texture recognition results with different models and different lengths of input sequences. }
		\label{tab:re1}
        \scalebox{0.8}{
		\begin{tabular}{c| c | c | c | c| c| c}
			\hline
			Models & $n=2$ & $n=3$ & $n=4$& $n=5$ & $n=6$& $n=7$ \\
			\hline
		CNNs & 67.23\% & 72.04\% & 75.26\% & 78.06\% & 79.56\% & 81.29\%\\
		CNNs+Spatial Attention & {72.12\%} & {73.97\%} & {78.60\%} & {80.43\%}  & {80.43\%}& {80.86\%}\\
		STAM & \pmb{76.50\%} & \pmb{79.35\%} & \pmb{80.00\%} & \pmb{80.64\%}  & \pmb{81.72\%}& \pmb{81.93\%}\\
		\hline
		\end{tabular}}
\end{table*}

\begin{table*}[htbp]
	\centering
		\caption{Texture recognition results while the dataset includes some noisy data collected before contact.}
		\label{tab:re2}
        \scalebox{0.8}{
		\begin{tabular}{c| c | c | c | c| c| c}
			\hline
			Models & $n=2$ & $n=3$ & $n=4$& $n=5$ & $n=6$& $n=7$ \\
			\hline
		CNNs & 53.20\% & 58.20\% & 59.60\% & 61.23\% & 64.60\% & 69.40\%\\
		CNNs+Spatial Attention & {55.40\%} & {60.80\%} & {62.60\%} & {62.80\%}  & {65.40\%}& {71.00\%}\\
		STAM & \pmb{72.00\%} & \pmb{72.20\%} & \pmb{75.80\%} & \pmb{76.61\%}  & \pmb{80.80\%}& \pmb{80.20\%}\\
		\hline
		\end{tabular}}
\end{table*}

\subsection{Experimental Results}\label{10}
The experiments mainly consist of three parts. 
1) An ablation study is performed that the spatial attention module is used separately at first, and then spatial attention and  temporal attention module are applied jointly to understand the effectiveness of the designed method step by step.
2) Short and long length of the sequence have a different effect on the the recognition performance and processing time. Therefore different length of input sequences is explored in our experiment.
3) Due to the uncertainty in the dynamic environment, some tactile images are collected before the contact happened. These tactile images are introduced as noisy data to test the robustness of the model.

As is shown in Table \ref{tab:re1} and Table \ref{tab:re2}, the proposed method has the best recognition results in all conditions with different lengths of sequences used.
Compared with using CNNs only, there is a $4.45\%$ average improvement in recognition accuracy by using proposed method.
In a further step, after introducing the noisy data, the accuracy increases by $15.23\%$ on average compared with baseline.
What is more, with a longer length of sequences applied, the accuracy has an upward trend.

From Table \ref{tab:re1}, it can be learnt that the lengths of sequences lead to differing impacts on performance for different models.
For example, when the lengths of sequences change from 7 to 2, the recognition accuracy of proposed methods only decrease $5.43\%$, the CNNs method and spatial attention method decrease $14.06\%$ and $8.74\%$ respectively. 
It can be found that, by using the spatial attention methods, the accuracy improves slightly in most cases compared with baseline.
By using both spatial attention and temporal attention, the performance has a further improvement, achieving the highest accuracy in all cases.
The results indicate that the proposed method has the ability to extract the obvious features effectively, especially with limited information.

From Table \ref{tab:re2}, it can be found that the introduction of noisy data leads to a lower accuracy on a whole for the texture recognition task.
However, the proposed method sustains a similar recognition results with a small decrease compared with Table \ref{tab:re1} while the other two methods have an obvious drop on the performance.
Specifically, when the length of sequence is 6, accuracy of baseline drop $14.96\%$ and the recognition accuracy of proposed method falls only $0.92\%$.
The results demonstrate that the proposed spatio-temporal attention method is robust to the noise and has the ability to select salient features effectively.

To summarise, two strengths of proposed model can be listed: 1) A recognition model with strong robustness is proposed. By using the attention mechanism, the informative information can be selected and noise can be restrained.
2) The learning of the feature is more efficient and effective.
It shows that the proposed method has a better performance under limited information when the length of sequence is short. 

\subsection{Attention Distribution Visualisation}\label{11}
In this section, the attention distributions are visualised for spatial attention and temporal attention respectively to illustrate how they work in the recognition tasks.

\begin{figure*}[t]
	\centering
	\includegraphics[width=1\columnwidth]{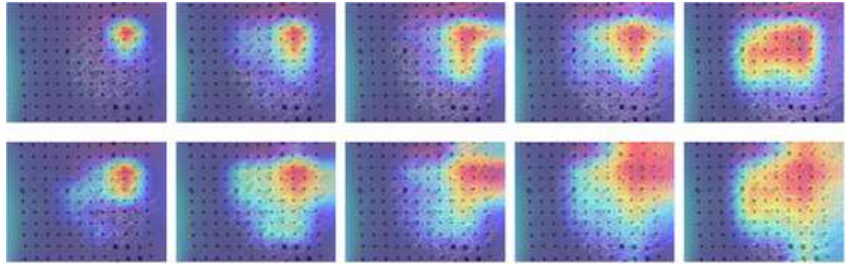}
	\caption{\textbf{\textit{Spatial Attention Distribution.}} 
	The first row and second row represent a same tactile sequence, but activated by non-attention method and spatial attention method with Grad-CAMs.
	It can be found that more contact regions are activated while using spatial attention method.} 
	\label{fig:heatmap}
\end{figure*}

\textit{\textbf{Spatial Attention Distribution.}} To visualise the spatial attention distribution, the gradient class activation maps (Grad-CAM) ~\cite{selvaraju2017grad} is performed on both spatial attention based methods and the methods using CNNs only. In the Grad-CAM, the gradients are used to express the saliency of region according to the classification results. 
The Grad-CAM allows us to understand the informative location in different methods.
As is demonstrated in Fig.\ref{fig:heatmap}, a larger contact regions are highlighted by the spatial attention method compared with baseline, which indicates that more informative locations are utilized by spatial attention.

\begin{figure*}[t]
	\centering
	\includegraphics[width=1\columnwidth]{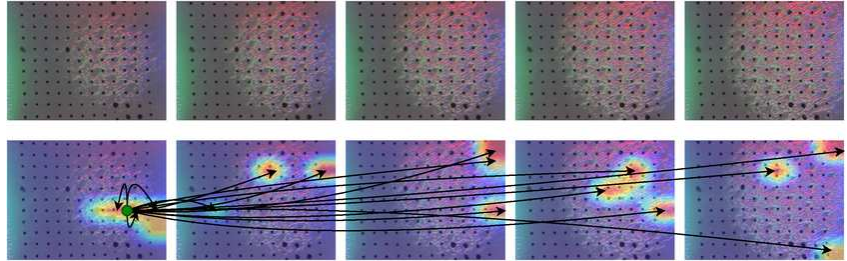}
	\caption{\textbf{\textit{Temporal Attention Distribution.}} Upper row: the sequence of tactile textures in order. Lower row: the temporal attention distribution where the green dot refers to a randomly selected regions, and the highlighted regions pointed by arrows are top 3 related regions in each texture. It can be found that most highlighted regions refer to latest contacted locations    } 
	\label{fig:selfdistribution}
\end{figure*}

\textit{\textbf{Temporal Attention Distribution.}} To understand the correlation between each locations regardless of time, the attention map of temporal attention is visualised. Due to the application of multiple temporal attention, the value of attention maps are averaged. As is shown in Fig.\ref{fig:selfdistribution}, a green dot is randomly selected in the first frame, then 3 most related locations which is pointed by arrow are highlighted according to the attention map in the other frames. As most highlighted regions are located in latest contact area, it shows that the temporal attention is able to make use of latest information in the time dimension for recognition task.

\label{sec:GelTip}

\section{Conclusion and Discussions}\label{7}
In this chapter, we discuss the multimodal perception for dexterous perception including the cross-modal learning, 3D reconstruction, multimodal translation as well as the attention mechanism. Due to the complementary characteristics between different modalities, it seems a more natural way to enable the robots multiple modalities to perceive the world as humans. As a result, the robots are able to observe and interact with the objects from different dimensions, multiplying the understanding of the physical world. 
Specifically, we discuss a cross-modal sensory data generation framework to make up inaccessible data, and  a spatio-temporal attention model for tactile texture recognition, which can learn the informative feature efficiently.
In future work, the selective attention mechanism can be used to facilitate the multimodal perception and manipulation.
There also exists many promising applications to be explored: 
1) Instead of fusing different modalities directly, the algorithms that can fuse the vision and tactile sense more effectively is worth to be investigated. 
2) Currently, most of the tasks are isolated learning paradigms. Hence, how to develop a lifelong learning framework to leverage the prior knowledge, improving the efficiency of the learning?
3) The robots always pays attention equally to the surroundings. How to enable the robots to act more like humans with selective attention, which is necessary to be explored.





\Backmatter
\begin{frontmatter}
 \chapter*{References}
 \markboth{References}{References}
 \bibliographystyle{elsarticle-num} 
 \bibliography{book}
\end{frontmatter}


\end{document}